\documentclass[journal]{IEEEtran}
\ifCLASSINFOpdf
\else
\fi
\usepackage{url}
\usepackage{graphicx}
\usepackage{amsfonts}
\usepackage{amsmath}
\usepackage{caption,subcaption}
\usepackage{cite}

\def\etal{\emph{et al}.}

\begin{document}
\urlstyle{tt}
%
\title{Towards Reading Beyond Faces for \\Sparsity-Aware 4D Affect Recognition}
%
%
%

\author{
		Muzammil Behzad, Nhat Vo, Xiaobai Li, and Guoying Zhao\\
		Center for Machine Vision and Signal Analysis (CMVS), University of Oulu, Finland.\\
		Email: \url{{muzammil.behzad, nhat.vo, xiaobai.li, guoying.zhao}@oulu.fi}


}

%
%

\markboth{IEEE Transactions on Multimedia,~Vol.~XX, No.~XX, August~2020}%
{Shell \MakeLowercase{\textit{et al.}}: Bare Demo of IEEEtran.cls for IEEE Journals}
%



\maketitle

\begin{abstract}
In this paper, we present a sparsity-aware deep network for automatic 4D facial expression recognition~(FER). Given 4D data, we first propose a novel augmentation method to combat the data limitation problem for deep learning. This is achieved by projecting the input data into RGB and depth map images and then iteratively performing randomized channel concatenation. Encoded in the given 3D landmarks, we also introduce an effective way to capture the facial muscle movements from three orthogonal plans (TOP), the TOP-landmarks over multi-views. Importantly, we then present a sparsity-aware deep network to compute the sparse representations of convolutional features over multi-views. This is not only effective for a higher recognition accuracy but is also computationally convenient. For training, the TOP-landmarks and sparse representations are used to train a long short-term memory (LSTM) network. The refined predictions are achieved when the learned features collaborate over multi-views. Extensive experimental results achieved on the BU-4DFE dataset show the significance of our method over the state-of-the-art methods by reaching a promising accuracy of 99.69$\%$ for 4D FER.
\end{abstract}


%
\IEEEpeerreviewmaketitle

\section{Introduction}
\label{sec:intro}

Facial expressions (FEs) are one of the most important, powerful and natural ways to convey and understand human emotions. Many works have been reported in this regard to automate facial expression analysis due to its tremendous potential in many applications \cite{fang20113d}, such as medical diagnosis, social robots, self-driving cars, educational well-being and several other human computer interaction methods. This paved the way for the rise of many facial expression recognition (FER) systems \cite{li2018deep} using the recently-trending computer vision technologies with deep learning. 

Particularly, FER is considered to create gateways for essential contributions to the next-generation interaction systems. However, FER brings along various challenging issues mainly due to the complexity of imaging environment and the inter-class diversity of facial expressions. Dating back to 1970s, the pioneer study done by Ekman and Friesen~\cite{ekman1971constants} discussed the six universal human facial expressions which are happiness, anger, sadness, fear, disgust and surprise. To recognize such expressions, many methods have been proposed in the literature. Depending upon the dimension of the input facial data, these methods can be loosely classified into the following four categories: (1) FER using static or dynamic 2D~images~\cite{fasel2003automatic,zeng2008survey,zheng2014multi, 7530823, 8371638}; (2) FER using 3D static face scans \cite{maalej2011shape, FANG2012738, 7163090, 6130397}; (3) FER using both 2D and 3D face data \cite{7944639,9115253, li2015efficient}; and (4) FER using 4D data or dynamic 3D face scans \cite{7457243, 6130440, behzad2019automatic, 8373807, 8023848}.

Despite the promising progress reported in the literature by 2D-based FER methods, they suffer from inevitable issues. Specifically, such methods face challenging problems of sensitivity towards pose variations, lighting condition changes and occlusions. With the advent of high-speed and high-resolution 3D data acquisition equipment, an essential alternative has been provided to achieve robust FER performance. This is due to the fact that 3D data additionally provides information about facial expressions in the form of deformation of various muscle movements on facial surfaces \cite{7457243}. They also offer immunity to the variations in viewpoint and lighting. As a result, the research direction is recently steered towards FER systems based on static and dynamic 3D face models.

More recently, the rapid development of imaging systems for 4D data acquisition has enabled the access to the dynamic flow of 3D shapes of improved quality for a detailed investigation of facial expressions. Importantly, being more robust towards the aforementioned problems, and providing additional geometry information \cite{amor20144}, the 4D data conveniently stores the facial deformations when a facial expression is triggered. In this regard, the release of large-size facial expression datasets containing 3D/4D face scans (e.g., the BU-4DFE dataset \cite{4813324}) has allowed 4D FER by fetching facial deformation patterns both spatially as well as temporally.

\subsection{Related Work}
\subsubsection{Advances in 3D Facial Expression Recognition}
To learn from the underlying 3D facial geometry, a number of methods are reported, which only rely on the static 3D data, i.e., apex frames. Generally, the most popular approaches can be categorized as local region or feature-based, template-based, 2D projections-based, and curve-based approaches

In local region or feature based techniques, different topological and geometric characteristics are extracted from several regions of a 3D face scans, and quantization of such characteristics is adopted to represent various expressions for better classification. Some typical examples are local normal patterns (LNP) \cite{6460694}, low level geometric feature pool \cite{5206613}, and histograms of differential geometry quantities \cite{li2015efficient}. On the other hand, template-based methods requires to fit deformable or rigid generic face model to an input face scan under certain criterion, and the computed coefficients are then regarded as extracted features of facial expressions. Some representative methods belonging to this category are annotated face model~\cite{6130397}, bi-linear models~\cite{4539275}, and statistic facial feature model~\cite{5597896}.

With an added advantage of reusing the conventional solutions for 2D FER, 2D projections-based approaches exploit various 2D representations from the given 3D face models. In such works, deep features are used for learning the facial attributes in 2D images projected from the 3D face models. For example, the authors in \cite{7944639} presented a 3D face model in terms of six types of projected 2D maps of facial attributes. These maps are then fed into a deep learning model for learning distinctive features along with fusion learning. Similarly, Oyedotun \etal \cite{8265585} proposed a deep learning model for joint learning of robust facial expression features from fused RGB and depth map latent representations.

In comparison with the above-mentioned methods, the curve-based approaches have also shown effective performance. By computing various shapes via a set of sampling curves in the Riemannian space, deformations caused by different facial expressions are represented. For instance, the authors in \cite{samir2009intrinsic} presented a framework for representing facial surfaces with the help of an indexed set of 3D closed curves. These curves correspond to level curves of a surface distance function which is defined as the shortest distance between a point of the computed curve on the facial surface and a reference nose tip point. Such representation allows comparing the 3D shapes by corresponding curves. Similarly in \cite{maalej2011shape}, the authors used the curve-based idea for 3D FER. They computed the distance of the geodesic path between the corresponding regions on a 3D face. This method produces quantitative information from the surfaces of various facial expressions which ultimately helps the classification performance.

\subsubsection{Towards 4D Facial Expression Recognition}
In the past decade, 4D FER has been widely investigated due to its significance towards better performance. This is mainly because 4D data provides complimentary spatio-temporal patterns allowing deep models to significantly understand and predict the facial expressions. For instance, Sun~\etal~\cite{Sun:2010:TVF:1820799.1820803} and Yin~\etal~\cite{4813324} proposed a method based on Hidden Markov Models (HMM) to learn the facial muscle patterns over time. In another attempt, using random forest, Ben Amor~\etal~\cite{amor20144} presented a deformation vector field mainly based on Riemannian analysis to benefit from local facial patterns. Likewise, Sandbach \etal~\cite{sandbach2012recognition} proposed free-form deformation as representations of 3D frames and then used HMM and GentleBoost for classification. Moreover, the authors in \cite{FANG2012738} used support vector machine (SVM) and represented geometrical coordinates and its normal as feature vectors, and as dynamic local binary patterns (LBP) in another work \cite{6130440}. Similarly, to extract features from polar angles and curvatures, a spatio-temporal LBP-based feature was proposed in \cite{6553746}.

By using the scattering operator \cite{bruna2013invariant} on 4D face scans, Yao~\etal~\cite{Yao:2018:TGS:3190503.3131345} applied multiple kernel learning (MKL) to produce effective feature representations. Similarly, the authors in \cite{fabiano2018spontaneous} proposed a statistical shape model with local and global constraints to recognize FEs. They claimed that local shape index and global face shape can be combined to build a required FER system. For automatic 4D FER via dynamic geometrical image network, an interesting model was proposed by Li \etal~\cite{8373807}. They generated geometrical images after the differential quantities were estimated where the final emotion prediction was a function of score-level fusion from different geometrical images. 

In a similar work \cite{behzad2019automatic}, a collaborative cross-domain dynamic image network was proposed for automatic 4D facial expression recognition. They computed geometrical images and combined their correlated information for a better recognition. Another recent method exploits the sparse coding-based representation of LBP difference \cite{Bejaoui2019}. The authors first extracted appearance and geometric features via mesh-local binary pattern difference (mesh-LBPD), and then applied sparse coding to recognize FEs.

\subsection{Motivation}
Despite many attempts to automate 4D FER, we believe there exist many loopholes that need attention. For example, the data available to train a deep network for 4D FER is very limited. This calls for efficient data augmentation techniques to satisfy the data-hungry nature for deep learning. Moreover, despite the fact that the multi-views of 4D faces jointly capture facial deformations, and that landmarks also encode specific movement patterns, their role is often ignored. Importantly, the need of appropriate facial feature representations is vital in the success of a 4D FER system.

\subsection{Contributions}
In the light of above discussion, our contributions in this work are 3-fold as follows:
\begin{itemize}
	\item We present $\infty$-augmentation which is a simple yet efficient method to combat 3D/4D data limitation problem.
	\item We introduce TOP-landmarks over multi-views to extract landmark cues from three orthogonal planes.
	\item We compute sparse representations from extracted deep features, which outperforms the state-of-the-art performance with reduced computational complexity.
\end{itemize}

The rest of the paper is organized as follows: our proposed 4D FER method is explained in Section \ref{sec:prop_method}. In Section \ref{sec:res}, we present and discuss our extensive experimental results to validate the efficiency of the proposed method. Finally, Section~\ref{sec:con} concludes the paper.

\section{Proposed Method}
\label{sec:prop_method}
In this section, we explain our proposed method for automatic 4D FER. First, we augment our data to extend the dataset size for better deep learning. Second, we extract the facial patterns encoded in landmarks on three orthogonal planes to aid our expression recognition system. We then extract deep features which are then collaboratively learned with landmarks for an accurate expression recognition. To leverage the correlations between $\infty$-augmentation, TOP-landmarks and our sparse representations, the proposed sparsity-aware deep learning is explained as follows.

\subsection{$\infty$-Augmentation}
\begin{figure}[t!]
	\centering
	\includegraphics[width=\linewidth]{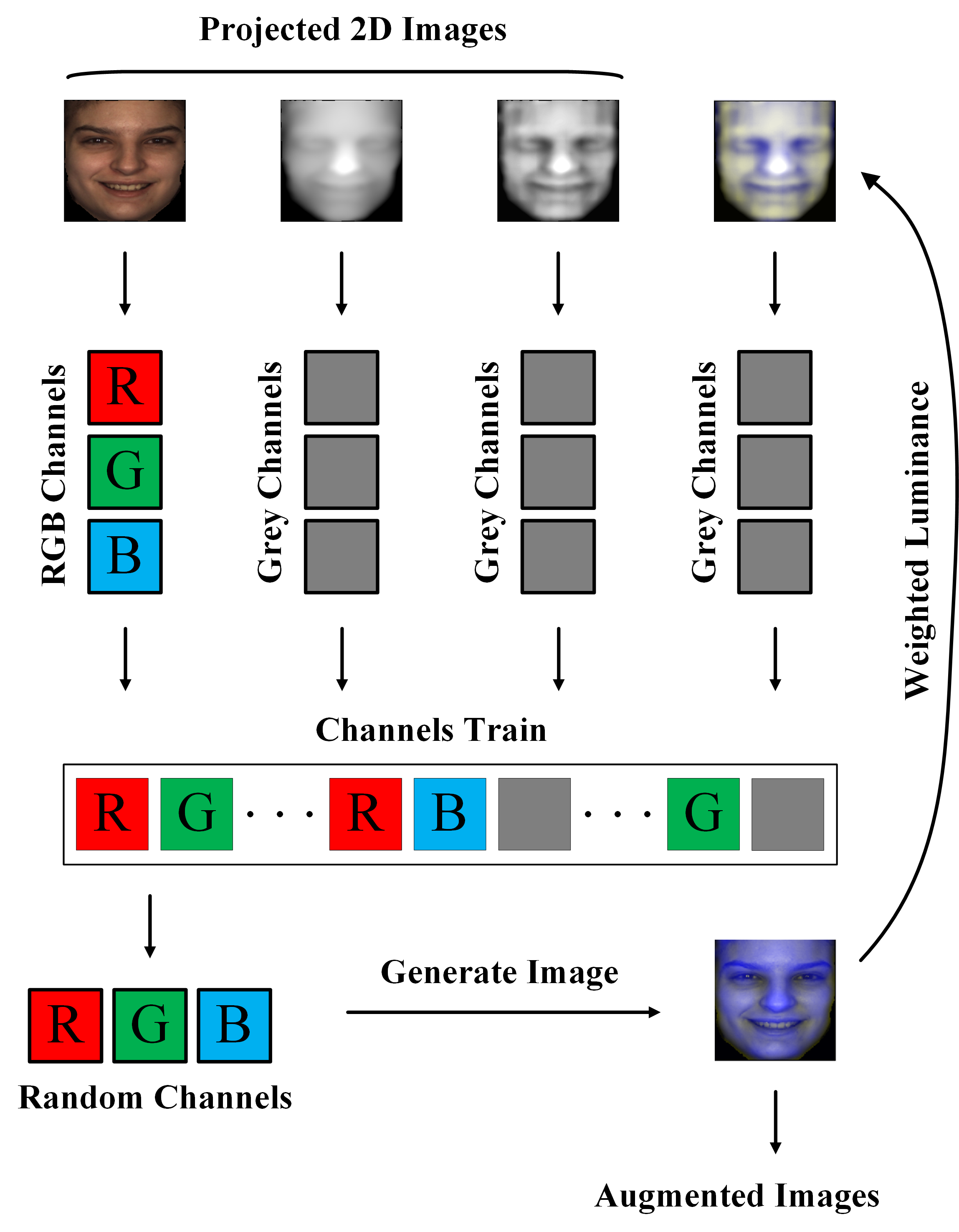}
	\caption{Working flow of the proposed $\infty$-augmentation method to generate new images.}
	\label{fig:4D_aug_method}
\end{figure}

To overcome the limitation of unavailability of large-scale 3D/4D dataset (e.g., only 606 videos in the BU-4DFE \cite{4813324} dataset), we propose a novel augmentation method. With our simple yet efficient method, the 3D/4D data can be theoretically augmented infinite times, hence the name, $\infty$-augmentation. For a given 4D dataset with $N$ examples, we process each 3D point-cloud independently. Therefore, we define
\begin{align}
\label{eq:1}
I^{4D} = \{I_{nt}^{3D}\},  \text{ } \forall t = \{1,2,3,...,T_n\} \text{ and }\forall n = \{1,2,3,...,N\},
\end{align}
where $I^{4D}$ is the set of 4D face scan examples, and $I_{nt}^{3D}$ denotes $n$th example and $t$th temporal frame. Note that (\ref{eq:1})~$\Rightarrow |I^{4D}| = N$, and $|I_{nt}^{3D}| = T_n$. Consequently for a given 3D point-cloud with $M$ vertices, let us denote its corresponding mesh as
\begin{align}
\label{eq:2}
\Psi = [\textbf{v}_1, ... ,\textbf{v}_M] = [(x_1,y_1,z_1),...,(x_M,y_M,z_M)],
\end{align}
where $\textbf{v}_j = [(x_j,y_j,z_j)]$ are the face-centered Cartesian coordinates of the $j$th vertex. Let us define $f_T: I^{3D} \rightarrow I_T$ to compute the projected RGB texture images, and $f_D: I^{3D} \rightarrow I_D$ to compute the projected depth images from the given mesh via 3D to 2D rendering, where $f_T$ and $f_D$ represent the function mapping from 3D mesh to texture image $I_T \in \mathbb{R}^{K^2}$, and depth image $I_D \in \mathbb{R}^{K^2}$, respectively, where $K^2$ is the number of pixels. For an image with richer details of the facial deformations, we apply contrast-limited adaptive histogram equalization on the depth images to get sharper depth images as $I_{ED} = \eta_s(I_D)$, where $\eta_s(.)$ represents the sharpening operator. Consequently, we get the three projected images as shown in Fig. \ref{fig:4D_aug_method}.

\begin{figure*}[t!]
	\centering
	\includegraphics[width=\linewidth]{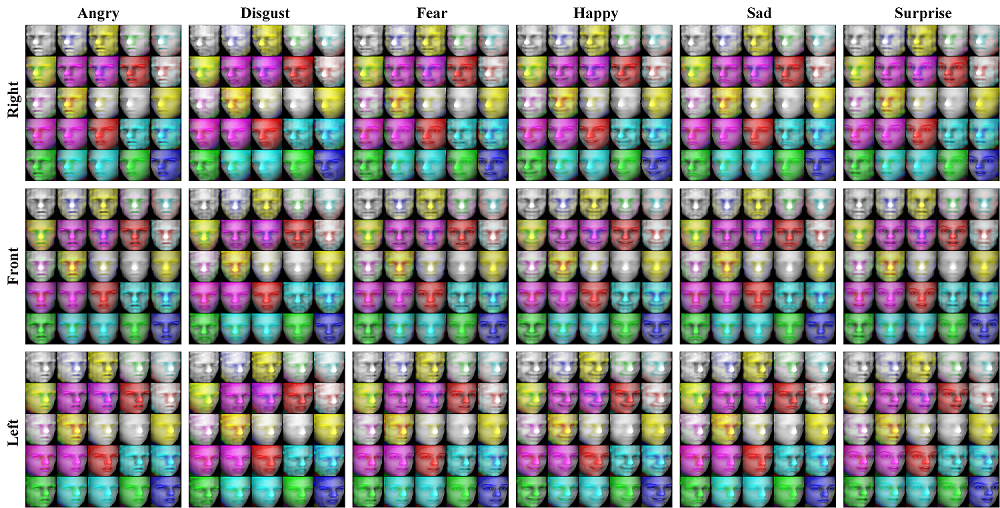}
	\caption{Augmented images generated using our proposed $\infty$-augmentation method for each of the six facial expressions. The rows represent the rotation angles at which the expressions are generated, while the columns correspond to the facial expressions. For an expression, we show three different profiles (right, front and left) and twenty five randomly selected samples to validate the effectiveness of our proposed augmentation method. The different colors help in visualizing how the underlying structure of the generated faces is different yet they exhibit similarities which fundamentally helps a deep learning network in identifying meaningful patterns.}
	\label{fig:4D_aug}
\end{figure*}
Once the projected 2D images are computed, the next step is to separate R/G/B and gray channel(s) of these images and stack them together to form a channels train. Afterwards, randomly selected channels from the channels train are concatenated to generate an augmented image $I_G$. Inspired by the RGB color model, we then propose to extract the luminance information $I_L$ from the selected channels using a weighted sum as
\begin{align}
\label{eq:3}
I_L = \alpha_1 I_G^R + \alpha_2 I_G^G + \alpha_3 I_G^B,
\end{align}
where $\alpha_i$ is the weight for each of the three R/G/B channels of the generated image $I_G$. Apart from the standard weights used for extracting luminance in converting a texture RGB image to a gray image ($\alpha_1=0.3,$~$\alpha_2=0.59,$~$\alpha_3=0.11,$), the flexibility in a wide range of weights allows us to perform $\infty$-augmentation conveniently. Importantly, by varying the order in which the selected channels are concatenated, and by including the weighted luminance information of an obtained augmented image as an input channel for the next rounds, we iteratively achieve $\infty$-augmentation. Note that this process is repeated for each 3D point-cloud or video frame.

In Fig. \ref{fig:4D_aug}, we show augmented images generated using our proposed $\infty$-augmentation method for each of the six universal expressions \footnote{For a detailed illustration of our proposed $\infty$-augmentation method, we suggest the readers to watch the video on YouTube provided in the following link for a better understanding of how the augmented images/videos appear in real time: \url{https://youtu.be/pnmzjpGLkb0}.}. The rows in this figure represent the rotation angles at which the expressions are generated, while the columns correspond to the facial expressions. For an expression, we show three different profiles (right, front and left) and twenty five randomly selected samples to validate the effectiveness of our proposed augmentation method. We use a rotation angle of 20 degrees to extract left and right profiles. In this figure, the different colors help in visualizing how the underlying structure of the generated faces is different yet they exhibit similarities which fundamentally helps a deep learning network in identifying meaningful patterns. Additionally, apart from the standard weights, the flexibility in choosing weights, the rotation angles and the order of channel concatenation leverages our proposed $\infty$-augmentation method and introduces new paradigms in the augmentation community.

\subsection{TOP-Landmarks}
\begin{figure}[b!]
	\centering
	\includegraphics[width=\linewidth]{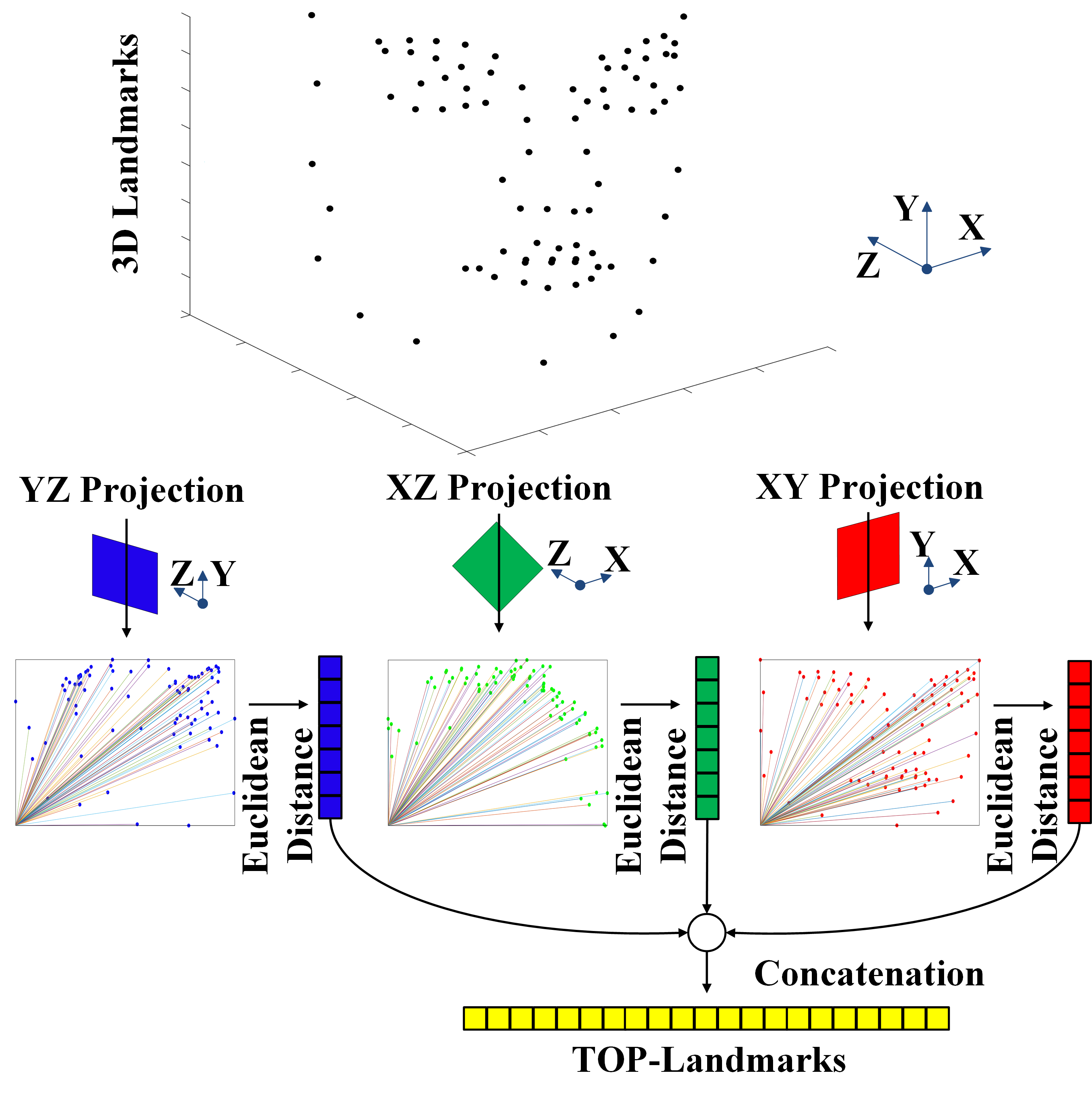}
	\caption{Calculation of TOP-landmarks.}
	\label{fig:TOPLandmakrs}
\end{figure}

Inspired by LBP-TOP \cite{4160945}, we propose TOP-landmarks to extract effective landmark cues from three orthogonal planes and then use it in our deep learning network over multi-views (left, front and right). For each of the given multi-views, we first project all the given 3D landmarks over the three XY, XZ and YZ orthogonal planes as shown in Fig.~\ref{fig:TOPLandmakrs}. Following definitions from (\ref{eq:2}), this means that for a given set of 3D landmarks with $m$ vertices, the corresponding projections in three orthogonal planes are 
\begin{align}
\label{eq:4}
\begin{split}
\psi_{XY} = [\textbf{v}_1, ... ,\textbf{v}_m] = [(x_1,y_1),...,(x_m,y_m)],\\
\psi_{XZ} = [\textbf{v}_1, ... ,\textbf{v}_m] = [(x_1,z_1),...,(x_m,z_m)],\\
\psi_{YZ} = [\textbf{v}_1, ... ,\textbf{v}_m] = [(y_1,z_1),...,(y_m,z_m)].
\end{split}
\end{align}
The projected points are then normalized to ensure correspondence across different frames. Next, we compute the Euclidean distance of each landmark point from its origin and store them as distance vectors. Finally, we concatenate the three distance vectors from each orthogonal plane to compute the resultant TOP-landmarks, denoted by $\Omega$, as
\begin{align}
\label{eq:5}
\Omega = \lVert \eta(\psi_{XY}) \rVert^\frown
\lVert \eta(\psi_{XZ}) \rVert^\frown
\lVert \eta(\psi_{YZ}) \rVert,
\end{align}
where $\eta(.)$ and ($^\frown$) denote the normalization and concatenation operators, respectively.

\begin{figure*}[t!]
	\centering
	\includegraphics[width=\linewidth]{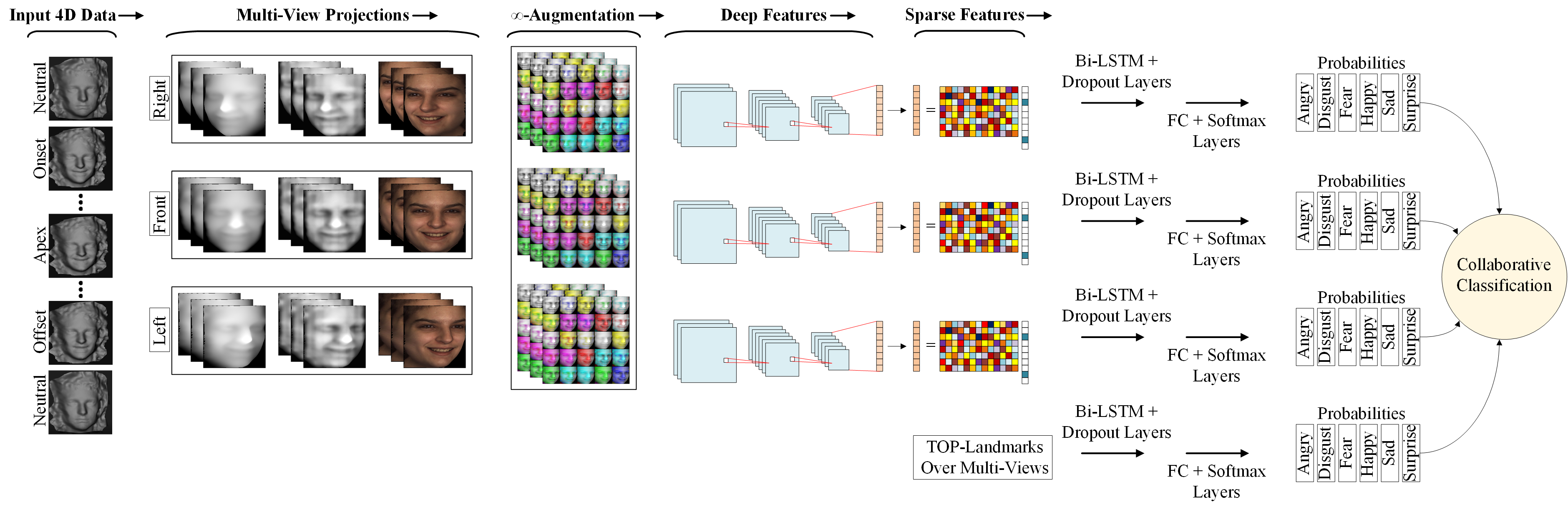}
	\caption{Overview of the proposed sparsity-aware deep learning for 4D affect recognition.}
	\label{fig:SADNet}
\end{figure*}
\subsection{Sparsity-Aware Deep Learning}
Sparsity has recently played a remarkable role in reducing the computation time and boosting classification performance in deep learning \cite{liu2015sparse}. Therefore, we also propose our deep network to be sparsity-aware as shown in Fig. \ref{fig:SADNet}. As shown in this figure, once we project the 4D data as RGB texture and depth maps over multi-views, we then augment the data for increasing the training data. Note that instead of using frontal view only, we resort to multi-views and then have a score-level fusion to incorporate the facial muscle movements from the side-views as well. For computational convenience, we use GoogLeNet \cite{szegedy2015going} to extract deep features from the augmented data. From a given 4D face scan, consider an input deep feature vector $\textbf{x}_k$ with length $P$, we aim to transform it into the equivalent sparse representations as
\begin{align}
\label{eq:sparse_equation}
\textbf{x}_k  = \textbf{A}\textbf{h}_k , \forall k
\end{align}
where $k$ is the index for current 3D face, $\textbf{A} \in \mathbb{R}^{P \times Q}, Q \gg P,$ is an overcomplete dictionary having wavelet basis. Furthermore, $\textbf{h}_k~\in~\mathbb{R}^{Q}$  is the equivalent sparse representation of the deep feature vector $\textbf{x}_k$. For the sparse reconstruction, we let $\widehat{\textbf{h}}_k$ denote an estimate of the sparse vector $\textbf{h}_k$ obtained via a sparse estimation algorithm \cite{blumensath2009iterative}, and let $\mathcal{S}_k$ represent the set of active indices in the sparse vector, i.e., its support set. For better estimates, each feature vector is processed individually. The estimate of sparse vector $\textbf{h}_k$ is computed as
\begin{align}
\label{eq:MMSE1}
\widehat{\textbf{h}}_k  = \mathop{\mathbb{E}}[\textbf{h}_k|\textbf{x}_k] =\sum_{\mathcal{S}_k} p(\mathcal{S}_k|\textbf{x}_k)\mathop{\mathbb{E}} [\textbf{h}_k|\textbf{x}_k,\mathcal{S}_k].
\end{align}
The following explains how the sum, the posterior $p(\mathcal{S}_k|\textbf{x}_k)$, and the expectation $\mathop{\mathbb{E}} [\textbf{h}_k|\textbf{x}_k,\mathcal{S}_k]$ in (\ref{eq:MMSE1}) are evaluated. Given the support~$\mathcal{S}_k$, (\ref{eq:sparse_equation}) becomes
\begin{align}\label{eq:MMSE2}
\textbf{x}_k = \textbf{A}_{\mathcal{S}_k}\textbf{h}_{\mathcal{S}_k},
\end{align}
where $\textbf{A}_{\mathcal{S}_k}$ is the matrix containing $\mathcal{S}_k$ indexed columns of $\textbf{A}$. Likewise, $\textbf{h}_{\mathcal{S}_k}$ is formed by $\mathcal{S}_k$ indexed entries of $\textbf{h}_k$. Since the distribution of $\textbf{h}_k$ is unknown making $\mathop{\mathbb{E}} [\textbf{h}_k|\textbf{x}_k,\mathcal{S}_k]$ very difficult to compute, we use the best linear unbiased estimate (BLUE) as
\begin{align}\label{eq:MMSE3}
\mathop{\mathbb{E}} [\textbf{h}_k|\textbf{x}_k,\mathcal{S}_k]	\gets	(\textbf{A}_{\mathcal{S}_k}^{H}\textbf{A}_{\mathcal{S}_k})^{-1}\textbf{A}_{\mathcal{S}_k}^{H}\textbf{x}_k,
\end{align}
where $(.)^H$ defines the Hermitian conjugate operator. Using Bayes rule, we can write the posterior as
\begin{align}\label{eq:MMSE4}
p(\mathcal{S}_k|\textbf{x}_k)=\frac{p(\textbf{x}_k|\mathcal{S}_k)p(\mathcal{S}_k)}{p(\textbf{x}_k)}.
\end{align}
We can ignore $p(\textbf{x}_k)$ in (\ref{eq:MMSE4}) as it's a common factor to all posterior probabilities. Since entries of $\textbf{h}_k$ are activated with Bernoulli distribution having $p$ as success probability, then
\begin{align}\label{eq:MMSE5}
p(\mathcal{S}_k)=p^{|\mathcal{S}_k|}(1-p)^{Q-|\mathcal{S}_k|}.
\end{align}
For $p(\textbf{x}_k|\mathcal{S}_k)$, if $\textbf{h}_{\mathcal{S}_k}$ is Gaussian, then $p(\textbf{x}_k|\mathcal{S}_k)$ would also be Gaussian which is easy to compute. On the contrary, evaluating $p(\textbf{x}_k|\mathcal{S}_k)$ is difficult for unknown or non-Gaussian $\textbf{h}_{\mathcal{S}_k}$ distribution. To tackle this, it can be noted that $\textbf{x}_k$ is formed by a vector in the subspace that is spanned by $\textbf{A}_{\mathcal{S}_k}$ columns. By projecting $\textbf{x}_k$ on orthogonal complement space of $\textbf{A}_{\mathcal{S}_k}$, the distribution can be computed. This is achieved using the projection matrix 
\begin{align}\label{eq:MMSE5a}
\textbf{Z}^{\perp}_{\mathcal{S}_k}=\textbf{I}-\textbf{Z}_{\mathcal{S}_k}=\textbf{I}-\textbf{A}_{\mathcal{S}_k}(\textbf{A}_{\mathcal{S}_k}^{H}\textbf{A}_{\mathcal{S}_k})^{-1}\textbf{A}_{\mathcal{S}_k}^{H}.
\end{align}
Dropping some exponential terms, and simplifying gives us
\begin{align}\label{eq:MMSE6}
p(\textbf{x}_k|\mathcal{S}_k)	\simeq \exp(\|\textbf{Z}^{\perp}_{\mathcal{S}_k}\textbf{x}_k\|^{2}).
\end{align}
In this way, we can evaluate the sum in (\ref{eq:MMSE1}) and the sparse representations are computed conveniently as depicted in Fig. \ref{fig:SADNet}. This is not only useful for accurate FER due to effective sparse representations, but is also computationally convenient by resizing deep features to fewer samples. For experimental convenience, we use sparse features with thirty samples. Note that our pre-determined dictionary contains wavelet basis which are suitable for the deep features as also validated by the results in the next section.

Finally, we create a long short-term memory (LSTM) network with a sequence input layer, Bi-LSTM layer with 2000 hidden units, 50$\%$ dropout layer followed by fully connected (FC), Softmax and classification layer. Consequently, we use the extracted sequences of TOP-landmark and sparse features, all over multi-views, to first train the LSTM network. Afterwards, score-level fusion is performed to collaboratively recognize expressions.

\section{Experimental Results}
\label{sec:res}
\subsection{Dataset and Experimental Settings}
To validate the efficiency of our proposed method, we use the BU-4DFE \cite{4813324} dataset. This dataset contains video clips of 58 females and 43 males (in total 101 subjects) having all six human facial expressions. Each clip has a frame rate of 25 frames per second (fps) lasting approximately 3 to 4 seconds. For experimental settings, we use a 10-fold subject-independent cross-validation (10-CV). Instead of using key-frames~\cite{Yao:2018:TGS:3190503.3131345} or employing sliding windows \cite{sandbach2012recognition}, we use entire 3D sequences. For extracting deep features, we use the pre-trained GoogLeNet \cite{szegedy2015going}. However, we also show our results on other pre-trained models. All of our experiments are carried out on a GP100GL GPU (Tesla P100-PCIE), and the training time takes approximately 3 days. Additionally, unless otherwise stated, no augmentation and three views (right, front and left) are used for all the experiments.
\begin{table}[b!]
	\centering
	\caption{Evaluation of the proposed $\infty$-Augmentation method on the BU-4DFE dataset for 4D facial expression recognition (FER).}
	\label{table:augmentation_comparison}
	\begin{center}
		\begin{tabular}{|l|c|}
			\hline
			Augmentation Size & FER Accuracy (\%) \\
			\hline\hline
			No Augmentation 			& 88.45 \\ \hline
			Augmentation with 5$\times$ samples	& 91.97 \\ \hline
			Augmentation with 15$\times$ samples   & 93.05 \\ \hline
			Augmentation with 25$\times$ samples 	& 93.70 \\ \hline
		\end{tabular}
	\end{center}
\end{table}

\subsection{Effect of $\infty$-Augmentation}
We evaluate our proposed $\infty$-augmentation method over different samples to prove its effectiveness towards deep networks. The accuracies of recognizing facial expressions in the BU-4DFE dataset are compared when different number of samples are generated while augmenting the data. We used random weights for each of the three channels in weighted luminance since the choice of weights is very flexible from~0~to~1. Importantly, to show the effectiveness of our augmentation method, we only do recognition over multi-views, and skip using TOP-landmarks or sparse representations. As shown in Table \ref{table:augmentation_comparison}, the results indicate an accuracy jump of 3.52\% when augmentation with 5$\times$ samples is used. This is related to the effective augmentation strategy where the underlying facial patterns are captured and restored to exhibit similar fundamental patterns that could help a deep network learn efficiently. This is worth mentioning again that we use a 10-fold subject-independent cross-validation to validate our experiments. This proves that our augmentation method is robust to over-fitting and makes it capable of performing effectively better in generalized experimental settings. Note that we have used a maximum of 25$\times$ augmented samples for convenience. More samples can be chosen at the cost of increased training time.

\subsection{Importance of Sparse Representations and TOP-Landmarks}
\begin{figure*}[t!]
	\begin{subfigure}{.46\linewidth}
		\centering
		\includegraphics[width=\textwidth]{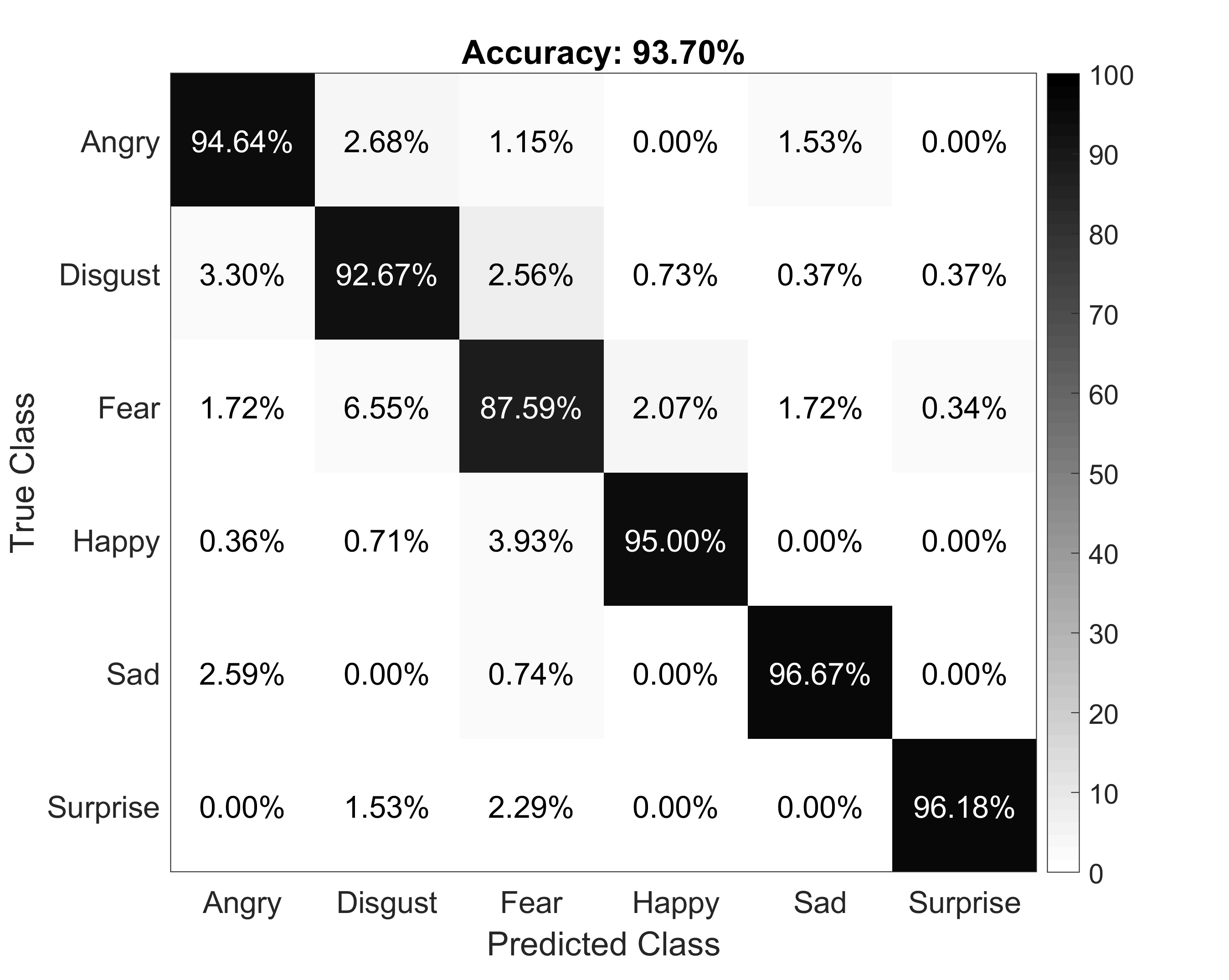}
		\caption{Dense}
		\label{fig:MultiviewsICIPA}
	\end{subfigure}\hspace{1cm}
	\begin{subfigure}{.46\linewidth}
		\centering
		\includegraphics[width=\textwidth]{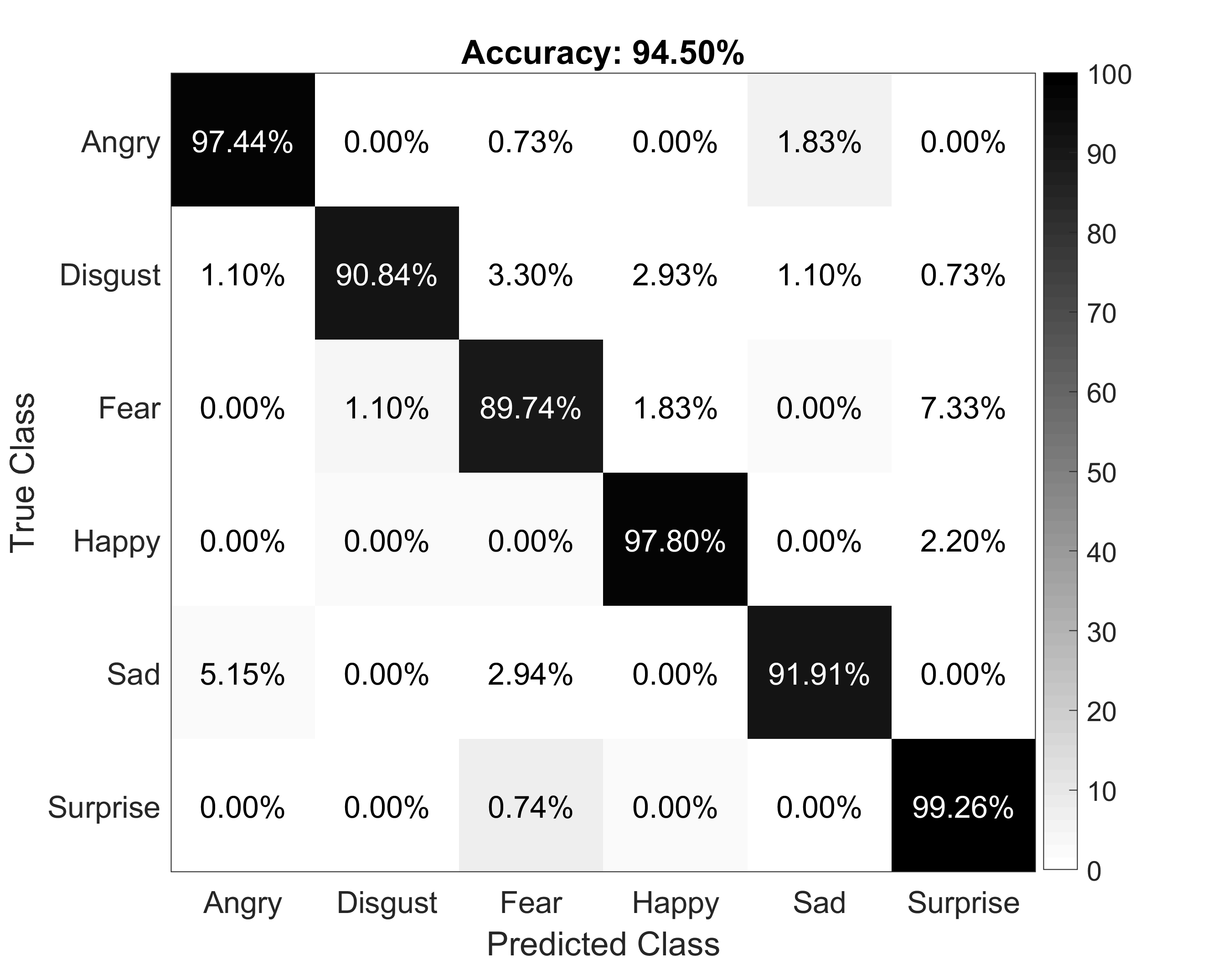}
		\caption{Sparse}
		\label{fig:MultiviewsICIPB}
	\end{subfigure}\\ \vspace{1.3cm}
	\begin{subfigure}{.46\linewidth}
		\centering
		\includegraphics[width=\textwidth]{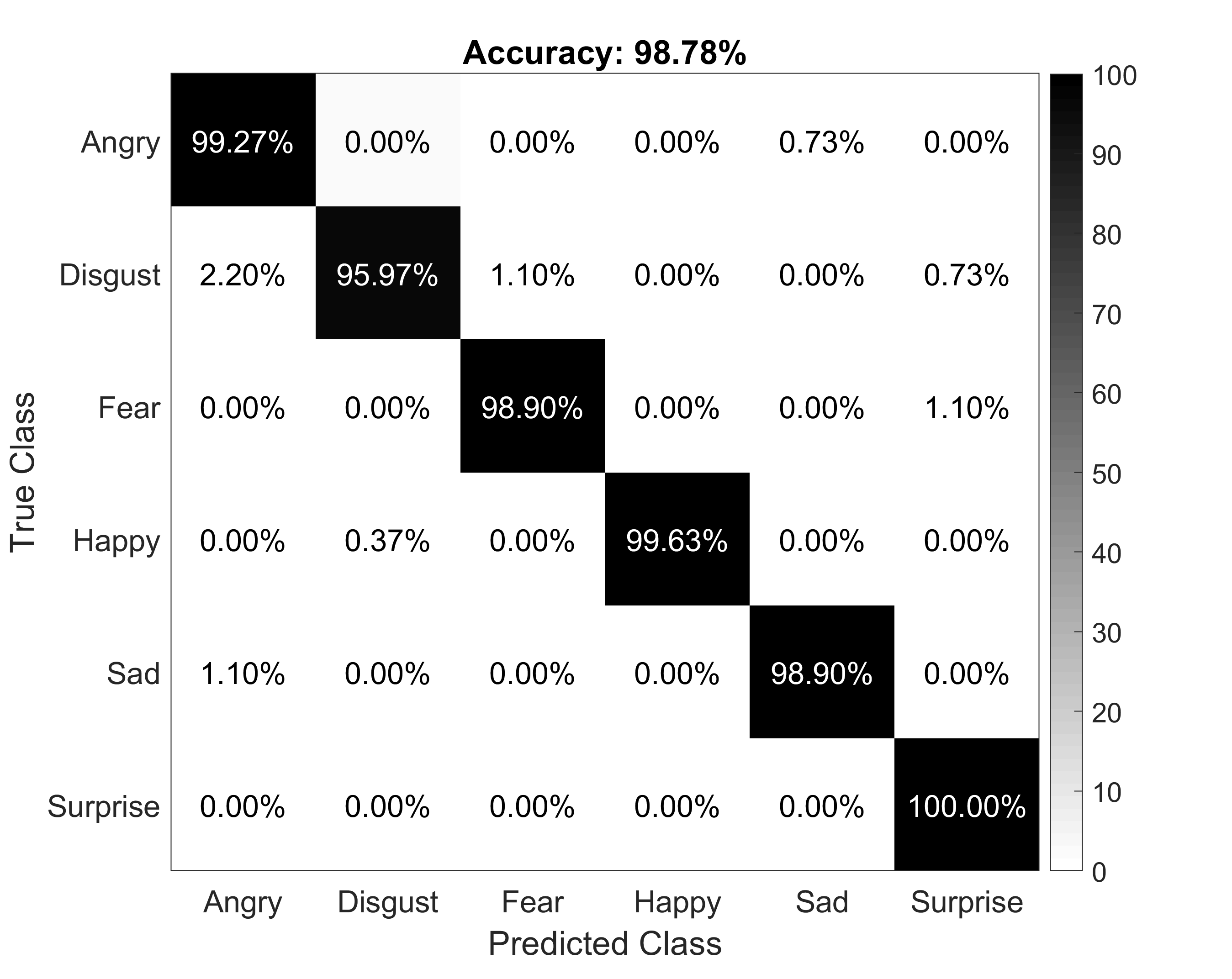}
		\caption{Dense + TOP-L}
		\label{fig:MultiviewsICIPC}
	\end{subfigure}\hspace{1cm}
	\begin{subfigure}{.46\linewidth}
		\centering
		\includegraphics[width=\textwidth]{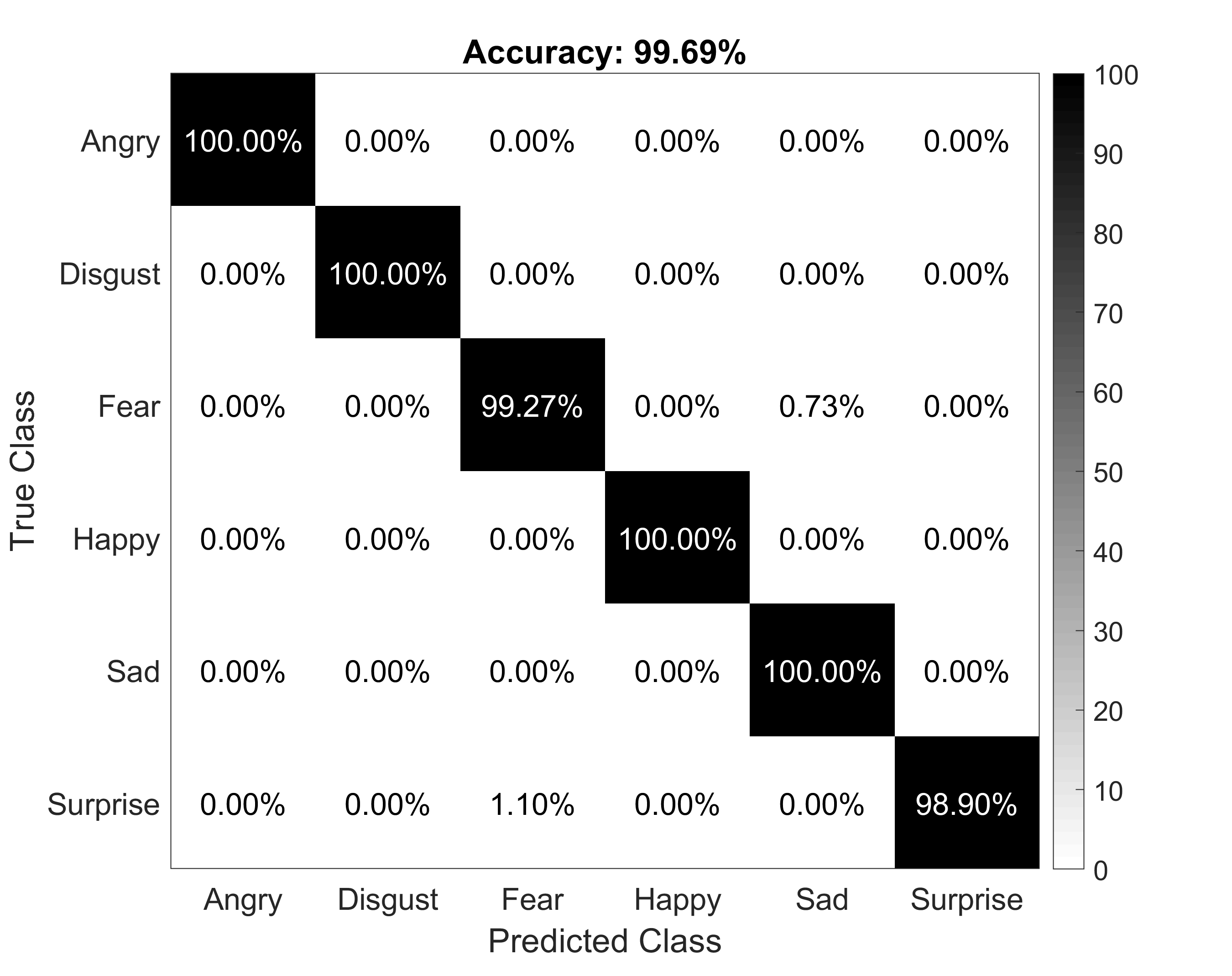}
		\caption{Sparse + TOP-L}
		\label{fig:MultiviewsICIPD}
	\end{subfigure}
	\caption{Comparison of confusion matrices over multi-views for 4D facial expression recognition. [TOP-L = TOP-landmarks]}
	\label{fig:MultiviewsICIP}
\end{figure*}

To highlight the importance of sparse representations and TOP-landmarks, we show the confusion matrices of our experiments in Fig. \ref{fig:MultiviewsICIP}. As shown in Fig. \ref{fig:MultiviewsICIPA}, when we use the dense representations only, an accuracy of 93.70\% is achieved. In comparison, the upshot of using sparse representations  instead of dense representations is an increased accuracy of 94.50\%, as shown in Fig. \ref{fig:MultiviewsICIPB}. An intuitively added benefit of the sparsity-aware learning is also the smaller computation time needed to process fewer samples of the sparse representations. A similar trend can be seen while using the TOP-landmarks. For instance, as shown in Fig. \ref{fig:MultiviewsICIPC}, a promising accuracy of 98.78\% is achieved when a joint recognition is performed with the help of effective cues computed via TOP-landmarks. It can be analyzed from this figure that despite stronger similarities between angry, disgust and fear expressions, the results indicate that our method accurately predicts the expressions validating its effectiveness. Finally, as shown in Fig. \ref{fig:MultiviewsICIPD}, a much higher accuracy of 99.69\% is achieved when employing sparsity-aware deep learning and TOP-landmarks over multi-views.

\subsection{Role of Multi-Views}
\begin{figure}[b!]
	\centering
	\includegraphics[width=\linewidth]{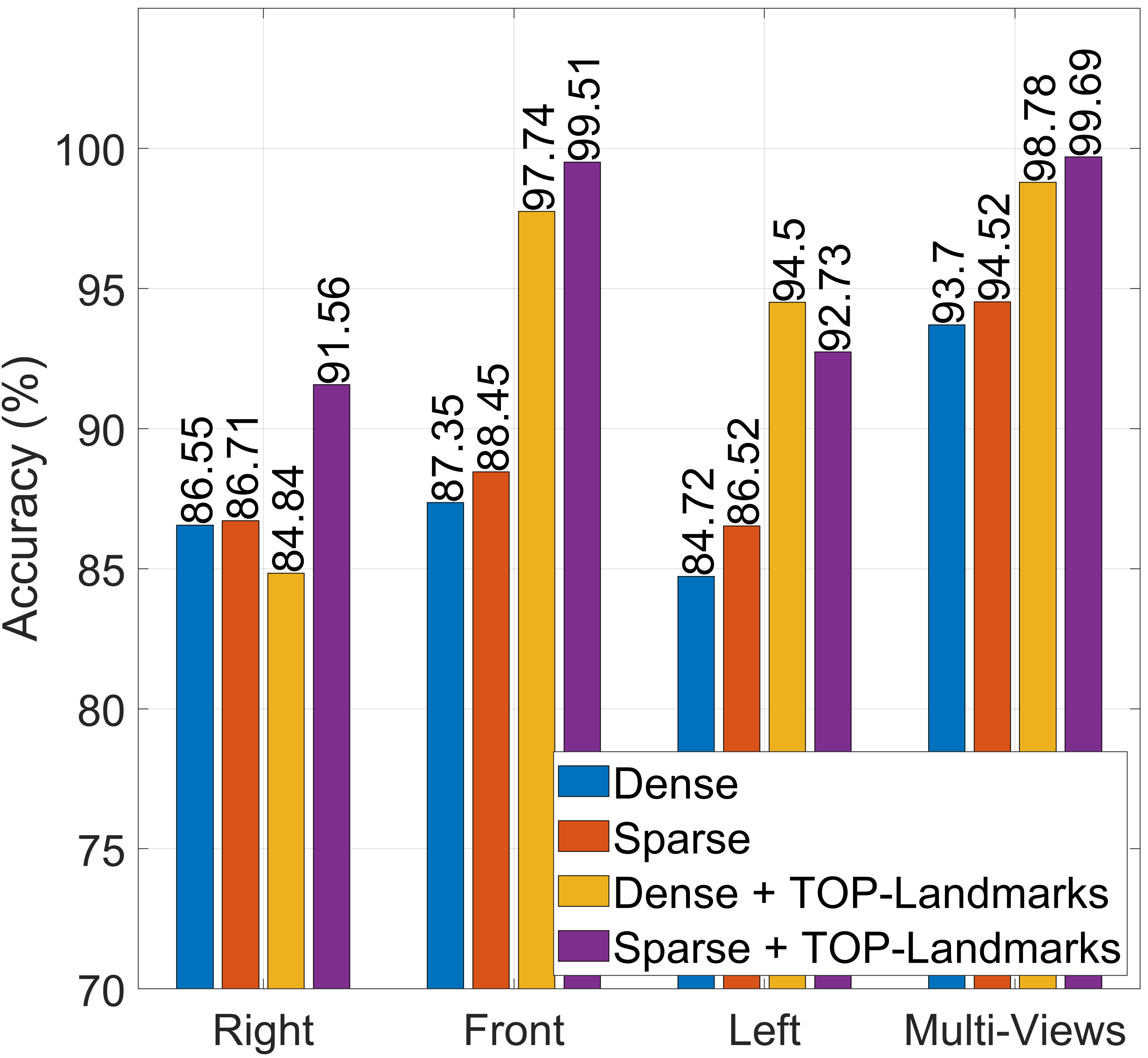}
	\caption{Accuracy ($\%$) comparison over multi-views for 4D facial expression recognition on the BU-4DFE dataset.}
	\label{fig:Multiviews}
\end{figure}

The results depicted in Fig.~\ref{fig:Multiviews} compare the contribution and effect of each of the multi-views. These results dictate that our method performs better when all the views are taken into account. For example, when we use only the dense representations over multi-views in our model, a promising accuracy of 93.70\% is still achieved which shows the effectiveness of the multi-views. Importantly, although the frontal view is more effective in comparison with left and right views, more effective results are achieved when a joint recognition is performed with the help of all these multi-views. More importantly, as discussed previously, a higher accuracy is achieved when both TOP-landmarks and sparse representations are utilized.
\begin{table}[b!]
	\caption{Accuracy ($\%$) comparison of 4D facial expression recognition with the state-of-the-art methods on the BU-4DFE dataset. [TOP-L = TOP-landmarks]}
	\label{table:4DFERresultsICIP}
	\begin{center}
		\begin{tabular}{|l|l|c|}
			\hline
			Method & Experimental Settings & Accuracy (\%) \\
			\hline\hline
			Sandbach \etal \cite{sandbach2012recognition} & 6-CV, Sliding window & 64.60 \\ \hline
			Fang \etal \cite{6130440} & 10-CV, Full sequence & 75.82 \\ \hline
			Xue \etal \cite{7045888} & 10-CV, Full sequence & 78.80 \\ \hline
			Sun \etal \cite{Sun:2010:TVF:1820799.1820803} & 10-CV, - & 83.70 \\ \hline
			Zhen \etal \cite{7457243} & 10-CV, Full sequence & 87.06 \\ \hline
			Yao \etal \cite{Yao:2018:TGS:3190503.3131345} & 10-CV, Key-frame & 87.61 \\ \hline
			Fang \etal \cite{FANG2012738} & 10-CV, - & 91.00 \\ \hline
			Li \etal \cite{8373807} & 10-CV, Full sequence & 92.22 \\ \hline
			Ben Amor \etal \cite{amor20144} & 10-CV, Full sequence & 93.21 \\ \hline
			Zhen \etal \cite{8023848} & 10-CV, Full sequence & 94.18 \\ \hline
			Bejaoui \etal \cite{Bejaoui2019} & 10-CV, Full sequence & 94.20 \\ \hline
			Zhen \etal \cite{8023848} & 10-CV, Key-frame & 95.13 \\ \hline
			Behzad \etal \cite{behzad2019automatic}  & 10-CV, Full sequence & 96.50 \\ \hline\hline
			\textbf{Ours (Dense)} & 10-CV, Full sequence & \textbf{93.70}\\ \hline
			\textbf{Ours (Sparse)} & 10-CV, Full sequence & \textbf{94.50}\\ \hline
			\textbf{Ours (Dense+TOP-L)} & 10-CV, Full sequence & \textbf{98.78}\\ \hline
			\textbf{Ours (Sparse+TOP-L)} & 10-CV, Full sequence & \textbf{99.69}\\ \hline
		\end{tabular}
	\end{center}
\end{table}

\subsection{Comparison with the State-of-the-Art Methods}
In Table \ref{table:4DFERresultsICIP}, we compare the accuracy of our method with several state-of-the-art methods \cite{7457243,Sun:2010:TVF:1820799.1820803,amor20144, sandbach2012recognition, FANG2012738, 6130440,Yao:2018:TGS:3190503.3131345,8373807, 7045888, 8023848, behzad2019automatic} on the BU-4DFE dataset. As illustrated, our method outperforms the existing methods in terms of correct expression recognition. This is mainly due to the extensively collaborative scheme of our method for an accurate expression recognition where the prediction scores are refined from its collaborators. Importantly, we show the effect of using dense vs. sparse representations and also how the TOP-landmarks assist in achieving substantial improvement. As shown, the sparse representations not only intuitively reduce the computational burden but also lead to a more accurate system by less over-fitting and better generalization, hence, raising the accuracy from $93.70\%$ to $98.78\%$. By using TOP-landmarks, our method further reaches a promising accuracy of $99.69\%$.

\subsection{Comparison with Other Pre-Trained Models}
Finally in Table \ref{table:OtherICIP}, we compare the results achieved by extracting deep features from other pre-trained models to validate the effectiveness of our method. Specifically, we use AlexNet~\cite{krizhevsky2012imagenet}, VGG16 \cite{russakovsky2015imagenet}, VGG19 \cite{russakovsky2015imagenet}, Inception-v3 \cite{szegedy2016rethinking}, ResNet18 \cite{he2016deep}, ResNet50 \cite{he2016deep} and ResNet101 \cite{he2016deep}. The table gives an overall impression that TOP-landmarks significantly improve the recognition accuracy. It also shows a similar trend, as seen in Table~\ref{table:4DFERresultsICIP}, that sparse features lead to a better system performance. The superior performance of GoogLeNet is due to several very small convolutions in order to drastically reduce the number of parameters.

\section{Conclusions}
\label{sec:con}
We proposed sparsity-aware deep learning to automate 4D FER. We first combated the problem of data limitation for deep learning by introducing $\infty$-augmentation. This method uses the projected RGB and depth map images, and then proceed to randomly concatenate them in channels over an iterative process. Second, we explained the idea of TOP-landmarks to capture the encoded facial deformations stored in the 3D landmarks. TOP-landmarks store the facial features from three orthogonal planes by using a distance-based approach. Importantly, we presented our sparsity-aware deep network where the convolutional deep features are used to compute deep sparse features which are then used to train an LSTM network and recognize expressions collaboratively with TOP-landmarks. With a promising accuracy of $99.69\%$, our method outperformed the existing state-of-the-art 4D FER methods in terms of expression recognition accuracy.

\begin{table}[t!]
	\centering
	\caption{Accuracy ($\%$) comparison of 4D facial expression recognition on the BU-4DFE dataset by using other pre-trained deep models.}
	\label{table:OtherICIP}
	\begin{center}
		\begin{tabular}{|l|c|c|c|}
			\hline
			Pre-trained Model & Dense & Dense+TOP-L & Sparse+TOP-L \\
			\hline\hline
			AlexNet \cite{krizhevsky2012imagenet}	  & 67.18 & 82.98 & 88.55 \\ \hline
			Inception-v3 \cite{szegedy2016rethinking} & 80.01 & 82.91 & 89.35 \\ \hline
			VGG16 \cite{russakovsky2015imagenet}      & 71.45 & 85.12 & 89.97 \\ \hline
			ResNet18 \cite{he2016deep} 				  & 76.34 & 87.56 & 91.60 \\ \hline
			ResNet101 \cite{he2016deep}			 	  & 76.65 & 87.71 & 91.70 \\ \hline
			ResNet50 \cite{he2016deep}			 	  & 83.74 & 91.06 & 93.93 \\ \hline
			VGG19 \cite{russakovsky2015imagenet} 	  & 84.54 & 91.66 & 94.33 \\ \hline
			GoogLeNet \cite{szegedy2015going}		  & 93.70 & 98.78 & 99.69 \\ \hline
		\end{tabular}
	\end{center}
\end{table}


%

\section*{Acknowledgments}
This work was supported by Infotech Oulu, the Academy of Finland for project MiGA (grant 316765), project 6+E (grant 323287), and ICT 2023 project (grant 328115). As well, the authors wish to acknowledge CSC – IT Center for Science, Finland, for computational resources.

\ifCLASSOPTIONcaptionsoff
  \newpage
\fi



\bibliographystyle{IEEEtran}
\bibliography{references}
\end{document}